\documentclass[journal]{IEEEtran}
\usepackage{amsfonts}
\usepackage{algorithm}
\usepackage{algorithmicx}
\usepackage{algpseudocode}
\usepackage{multirow}

\usepackage{cite}

\ifCLASSINFOpdf
  \usepackage[pdftex]{graphicx}
\else
  \usepackage[dvips]{graphicx}
\fi
\usepackage[cmex10]{amsmath}
\usepackage{array}

\ifCLASSOPTIONcompsoc
  \usepackage[caption=false,font=normalsize,labelfont=sf,textfont=sf]{subfig}
\else
  \usepackage[caption=false,font=footnotesize]{subfig}
\fi
\usepackage{url}

\begin{document}
%
\title{Manifold Regularized Slow Feature Analysis for Dynamic Texture Recognition}
%
%
%

\author{Jie~Miao,
        Xiangmin~Xu,~\IEEEmembership{Member,~IEEE,}
        Xiaofen~Xing,
        and~Dacheng~Tao,~\IEEEmembership{Fellow,~IEEE}
\thanks{Xiangmin Xu is the corresponding author.}
\thanks{J. Miao, X. Xu and X. Xing are with 
 School of Electronic and Information Engineering, 
 South China University of Technology, 
 Wushan RD., Tianhe District, Guangzhou, P.R.China. 
 E-mail: (miaow1988@qq.com; xmxu@scut.edu.cn; xfxing@scut.edu.cn;).
 }
\thanks{D. Tao is with the Centre for Quantum Computation \& Intelligent Systems and the Faculty of Engineering and Information Technology, University of Technology, Sydney, 81 Broadway Street, Ultimo, NSW 2007, Australia.
 E-mail: dacheng.tao@uts.edu.au
 }
}

%
%

\markboth{}
{Shell \MakeLowercase{\textit{et al.}}: Bare Demo of IEEEtran.cls for Journals}
%



\maketitle

\begin{abstract}
Dynamic textures exist in various forms, e.g., fire, smoke, and traffic jams, but recognizing dynamic texture is challenging due to the complex temporal variations.
In this paper, we present a novel approach stemmed from slow feature analysis (SFA) for dynamic texture recognition. 
SFA extracts slowly varying features from fast varying signals.
Fortunately, SFA is capable to leach invariant representations from dynamic textures.
However, complex temporal variations require high-level semantic representations to fully achieve temporal slowness, and thus it is impractical to learn a high-level representation from dynamic textures directly by SFA. 
In order to learn a robust low-level feature to resolve the complexity of dynamic textures, we propose manifold regularized SFA (MR-SFA) by exploring the neighbor relationship of the initial state of each temporal transition and retaining the locality of their variations.
Therefore, the learned features are not only slowly varying, but also partly predictable. 
MR-SFA for dynamic texture recognition is proposed in the following steps: 1) learning feature extraction functions as convolution filters by MR-SFA, 2) extracting local features by convolution and pooling, and 3) employing Fisher vectors to form a video-level representation for classification.
Experimental results on dynamic texture and dynamic scene recognition datasets validate the effectiveness of the proposed approach.
\end{abstract}


%
\IEEEpeerreviewmaketitle

\section{Introduction}

Dynamic texture is an extension of texture into the temporal domain. Dynamic textures exist in the real world in various forms, e.g., fire, smoke, water, human crowds, and traffic jams. 
Dynamic texture recognition can be used for many applications, e.g., fire detection, traffic monitoring, scene recognition, facial expression recognition and age estimation. 
Static cues are not sufficient for dynamic texture recognition.
Dynamic texture is a complex temporal process that takes place in the pixel domain. 
Non-rigid deformations in dynamic textures make the application of traditional computer vision approaches very challenging.
For example, optical flow requires motion smoothness, and a histogram of gradients requires clear edges and boundaries. Neither of these conditions can be fulfilled by dynamic textures.

Although much effort has been made, dynamic texture recognition remains a challenging problem.
A linear dynamical systems (LDS) approach attempts to model dynamic textures by a statistical generative model \cite{LDS}. However, LDS is sensitive to viewpoints, scale, rotation, and other factors.
Some carefully designed hand-crafted features (e.g., local binary patterns \cite{LBP-TOP}) describe dynamic textures by capturing the appearances and temporal variations. They tend to suffer from complex temporal variations, for example, non-rigid deformations and spatial-temporal translations.
In contrast to these approaches, we attempt to resolve the temporal complexity of dynamic textures. 
Once the temporal complexity is untangled, dynamic textures can be represented well.

Intuitively, the complexity of dynamic textures requires temporally invariant representations.
Inspired by the temporal slowness principle, slow feature analysis (SFA) extracts slowly varying features from fast varying signals \cite{SFA}.
For example, pixels in a video of dynamic texture vary quickly over the short term, but the high-level semantic information of the video varies slowly over the long term.
Fortunately, SFA is capable to leach invariant representations from dynamic textures.
However, the complex temporal variations that exist in dynamic textures require high-level semantic representations, which cannot be obtained directly by SFA.
Kernel methods \cite{SFA_SparseKernel} and non-linear expansions \cite{SFA_Action} were employed to reduce the gap between high-dimensional fast varying inputs and slowly varying high-level semantic representations. However, they are still not sufficient to extract a robust representation for dynamic texture recognition.

To address temporal complexity in dynamic texture recognition, we learn slowly varying features for local video volumes, and then, we obtain video-level representations by bag-of-words models.
In this way, local video volumes are well represented by learned features, and the video-level representation is invariant to translations, viewpoints, scales, and other aspects.
We further improve the standard SFA by exploring the manifold regularization \cite{Manifold} to ensure that the learned features are not only slowly varying but also partly predictable.
Specifically, we construct a neighbor relationship of all temporal transitions by their initial states, and then constrain the locality of their variations in the learned feature space. 
Consequently, each temporal variation can be partly predicted by its initial state, and the temporal complexity in the dynamic textures can be resolved better.
The evaluation on dynamic texture and scene recognition datasets shows that competitive results can be achieved compared with state-of-the-art approaches.

The remainder of this paper is organized as follows. 
Section~\ref{sec_related_work} discusses related studies. 
Sections~\ref{sec_mrsfa} and \ref{sec_mrsfa_features} detail the proposed approach.
The experimental results are presented in Section~\ref{sec_experiments}, and the conclusions are drawn in Section~\ref{sec_conclusion}.

\section{Related Work}
\label{sec_related_work}

This section discusses related work on dynamic texture recognition, and briefly reviews slow feature analysis and its improvements.

\subsection{Dynamic Texture Recognition}

A linear dynamical systems (LDS) approach for dynamic texture recognition was proposed assuming that dynamic textures are stationary stochastic processes \cite{LDS}. 
LDS is a statistical generative model. It can be further used for dynamic texture synthesis \cite{Superresolution}.
The recognition is performed by comparing the parameters of LDS. 
Some kernel methods and distance learning approaches were then proposed to improve the comparison \cite{Traffic,DynTex++}; however, their results are still limited by LDS-based features, which cannot handle different viewpoints, scales, or other aspects.
A bag-of-words model based on LDS features was proposed to improve conventional LDS-based approaches \cite{BagLDS}. Then, the bag-of-system-trees was further proposed for better efficiency \cite{BagTrees}.
Extreme learning machine (ELM) was applied to construct the codebook of LDS features while preserving the spatial and temporal characteristics of dynamic textures \cite{ExtremeLearning}.
A hierarchical expectation maximization algorithm was proposed to cluster dynamic textures using LDS features \cite{HierarchicalEM}.
The mixture of LDS was also exploited for modeling, clustering and segmenting dynamic textures \cite{MixtureLDS}. 
Although LDS is reasonable and intuitive, it tends to suffer from complex temporal variations in the sequential process. 

Local features have been successfully applied to dynamic texture recognition. 
Local binary patterns on three orthogonal planes (LBP-TOP) were proposed for dynamic texture and facial expression recognition \cite{LBP-TOP}. Instead of processing the entire video, this approach extracts features from three orthogonal planes in the video cube. LBP-TOP has been generalized to the tensor orthogonal
LBP for micro-expression recognition \cite{MicroExpression}.
Similar to LBP-TOP, the method of multiscale binarized statistical image features on three orthogonal planes (MBSIF-TOP) was proposed using binarized responses of filters learned by applying independent component analysis on each plane \cite{MultiscaleBinarized}. 
By capturing the direction of natural flows, a spatiotemporal directional number transitional graph (DNG) was proposed using spatial structures and motions of each local region \cite{DirectionalNumber}.
Although these approaches work well, they neglect a large amount of spatial-temporal information.

Some approaches have been proposed to fully utilize the spatial-temporal information.
The spatio-temporal fractal analysis (DFS) was proposed using both volumetric and multi-slice dynamic fractal spectrum components \cite{FractalAnalysis}.
Space-time orientation distributions generated by 3D Gaussian derivative filters were used for dynamic texture recognition \cite{YUPENN,derpanis2012spacetime}, and they have been successfully extended to bag-of-words models for dynamic scene recognition \cite{BagsEnergies}. 
Although both space and time were considered, the performance of these approaches are affected by the complexity of spatial-temporal variations.
Recently, a high-order hidden Markov model was employed to model dynamic textures \cite{qiao2015hidden}.
A dynamic shape and appearance model was proposed by learning a statistical model of the variability directly by a Gauss-Markov model \cite{DynamicShapeAppearance}.
A motion estimation approach based on locally and globally varying models was proposed to estimate optical flows in dynamic texture videos \cite{MotionEstimation}.
Besides the pixel domain, a wavelet domain multi-fractal analysis for dynamic texture recognition was proposed, and good results can be achieved by simply using frame averages \cite{Wavelet}. 

High-level features have also been exploited for dynamic texture recognition.
Deep learning has been successfully applied to general object recognition and detection. It has also been applied to dynamic texture recognition.
A 3D convolutional neural network (CNN) was trained from a very large number of videos \cite{C3D}. This 3D CNN has been used as general video feature extractor, and achieved a good result on dynamic scene recognition.
Many approaches use a pre-trained CNN as a high-level feature extractor \cite{TransferringCNN,SA_CNN,BeyondGauss}. These approaches outperform most existing dynamic texture recognition approaches. Besides the CNN, a complex network was proposed to extract features from dynamic textures directly \cite{goncalves_complex_2015}. A deep belief network was used to extract features from conventional features \cite{wang_exploiting_2015}.
In contrast to all of the above-mentioned approaches that are based on deeply learned networks, MR-SFA extracts features without using deep networks.

\subsection{Slow Feature Analysis}

Slow feature analysis (SFA) was proposed as an unsupervised learning approach \cite{SFA}. Inspired by the temporal slowness principle, SFA extracts slowly varying features from fast varying signals.
It has been proven that the properties of feature extraction functions learned by SFA are similar to complex cells in the primary visual cortex (V1) of the brain \cite{SFA_Bio}.
SFA has been successfully applied to applications such as human action recognition \cite{SFA_Action,TVA}, dynamic scene recognition \cite{SFA_Scene}, and blind source separation \cite{SFA_Separation,ExtensionSFA}. 

It is impractical to apply SFA to an entire video, which is extremely high dimensional. 
A possible solution is to extract local features from a small receptive field and then, use them for subsequent processing.
Zhang and Tao \cite{SFA_Action} employed SFA and nonlinear expansion to learn slow features of local cubes and use their accumulation as the video representation. A discriminative SFA was also proposed in their work to further improve the recognition result. However, this approach cannot generalize well to complex videos due to its dependency on simple and clear foregrounds.
Some improvements have been proposed to handle complex videos \cite{SFA_DL,TVA}. 
Inspired by deep learning, a hierarchical approach based on SFA was proposed \cite{SFA_DL}. This approach effectively extends the receptive field by a two-layer SFA feature extraction framework, and models videos by bag-of-words models. 
Afterward, SFA was generalized to temporal variance analysis to utilize both slow and fast features \cite{TVA}. Although fast varying motion features outperform slowly varying appearance features, temporal variance analysis relies on stabilized local volumes that are tracked by optical flows. 
In contrast to videos of human action, non-rigid deformations in dynamic textures are more challenging. 
It is difficult to extract a robust slowly varying feature for dynamic textures directly by SFA.
To accomplish this goal, Theriault \textit{et al.} \cite{SFA_Scene} employed SFA as a post-processing of Gabor features for dynamic scene recognition. Although significant improvements can be achieved compared with conventional Gabor features, the result is far from good compared with other approaches.

Many other improvements in SFA have also been proposed. 
A regularized sparse kernel SFA was proposed to generate feature spaces for linear algorithms \cite{SFA_SparseKernel}.
A changing detection algorithm based on an online kernel SFA was proposed for video segmentation and tracking \cite{SFA_Tracking}.
Although kernel methods can handle nonlinear data, they will introduce more noises and computational complexities than linear approaches.
Minh and Wiskott \cite{SFA_Separation} proposed a multivariate SFA for blind source separation.
A probabilistic SFA was proposed for facial behavior analysis \cite{ProbabilisticSFA}. 
Slow feature discriminant analysis (SFDA) was proposed as a supervised learning approach by maximizing the inter-class temporal variance and minimizing the intra-class temporal variance simultaneously \cite{SFDA}.
These approaches cannot be applied to dynamic texture recognition directly.

\section{Manifold Regularized Slow Feature Analysis}
\label{sec_mrsfa}

This section describes mathematical details about the proposed manifold regularized SFA (MR-SFA).
Matrices, vectors and scalars are denoted by uppercase letters, boldface lowercase letters and regular lowercase letters respectively (e.g. matrix $X$, vector $\mathbf{x}$ and scalar $x$).
All of the vectors in the paper are column vectors. The matrix and vector transpose is denoted by the superscript T. For example, $X^T$ is the transpose of $X$.

\subsection{Slow Feature Analysis}

First, we give a brief introduction on slow feature analysis (SFA) \cite{SFA}.
SFA is an unsupervised learning approach that extracts slowly varying features from fast varying signals. Here, we consider only one temporal sequence for simplicity. 
We denote a temporal sequence as $X = [\mathbf{x}_1, \cdots, \mathbf{x}_t] \in \mathbb{R}^{p \times t}$, where $\mathbf{x}_i$ is the state at time $i$. 
Without loss of generality, we assume that the input sequence $\{ \mathbf{x}_i \}$ is centered, i.e., we have $\sum_{i=1}^t \mathbf{x}_i = \mathbf{0}$.
SFA learns a new representation $Y = [\mathbf{y}_1, \cdots, \mathbf{y}_t] \in \mathbb{R}^{q \times t}$ which globally minimizes the overall temporal variation of $X$. Defining the temporal variation at time $i$ as $\mathbf{\dot{y}}_{i}=\mathbf{y}_{i} - \mathbf{y}_{i+1}$, the objective function of SFA can be formulated as
\begin{equation}
	\arg \min_Y \sum_i^{t-1} ||\mathbf{\dot{y}}_i||_2^2 \indent \textit{s.t.} \indent YY^T=I,
\end{equation}
where $I$ is an identity matrix. The constraint $YY^T=I$ guarantees a nontrivial solution.
Considering the linear case that $Y$ is obtained by an affine function $Y=U^T X$, where $U \in \mathbb{R}^{p \times q}$, we have
\begin{equation}
	\sum_i^{t-1} ||\mathbf{\dot{y}}_i||_2^2 = tr(\dot{Y} \dot{Y}^T) = tr(U^T (\dot{X} \dot{X}^T) U),
\end{equation}
where $tr(\cdot)$ is the matrix trace operator, $\dot{Y} = [\mathbf{\dot{y}}_1, \cdots, \mathbf{\dot{y}}_{t-1}] \in \mathbb{R}^{q \times (t-1)}$ and $\dot{X} = [\mathbf{\dot{x}}_1, \cdots, \mathbf{\dot{x}}_{t-1}] \in \mathbb{R}^{p \times (t-1)}$.
For simplicity, we further assume that the input sequence $\{ \mathbf{x}_i \}$ is whitened. In particular, we have
\begin{equation}
	XX^T = I.
\end{equation}
Therefore, the constraint $YY^T = I$ can be simplified as 
\begin{equation}
	YY^T = U^T (XX^T) U = U^T U = I.
\end{equation}
Lastly, the objective function of SFA can be reformulated as
\begin{equation}
	\arg \min_U tr(U^T (\dot{X} \dot{X}^T) U) \indent \textit{s.t.} \indent U^TU=I,
\end{equation}
and the solution $U$ can be obtained by solving the eigen-decomposition problem
\begin{equation}
	(\dot{X} \dot{X}^T) U = \Lambda U, 
\end{equation}
where $\Lambda$ is a diagonal matrix of eigenvalues.

\subsection{Manifold Regularized Slow Feature Analysis}

Standard SFA simply minimizes the overall temporal variation.
Non-rigid deformations in dynamic textures result in complex and noisy temporal transitions. Although features learned by standard SFA are slowly varying, they contain a large amount of noise.
To improve standard SFA, we explore the manifold regularization \cite{Manifold} that is based on a simple intuition: temporal features should not only be slowly varying, but also be predictable.

\begin{figure}
	\centering
	\includegraphics[width=3.4in]{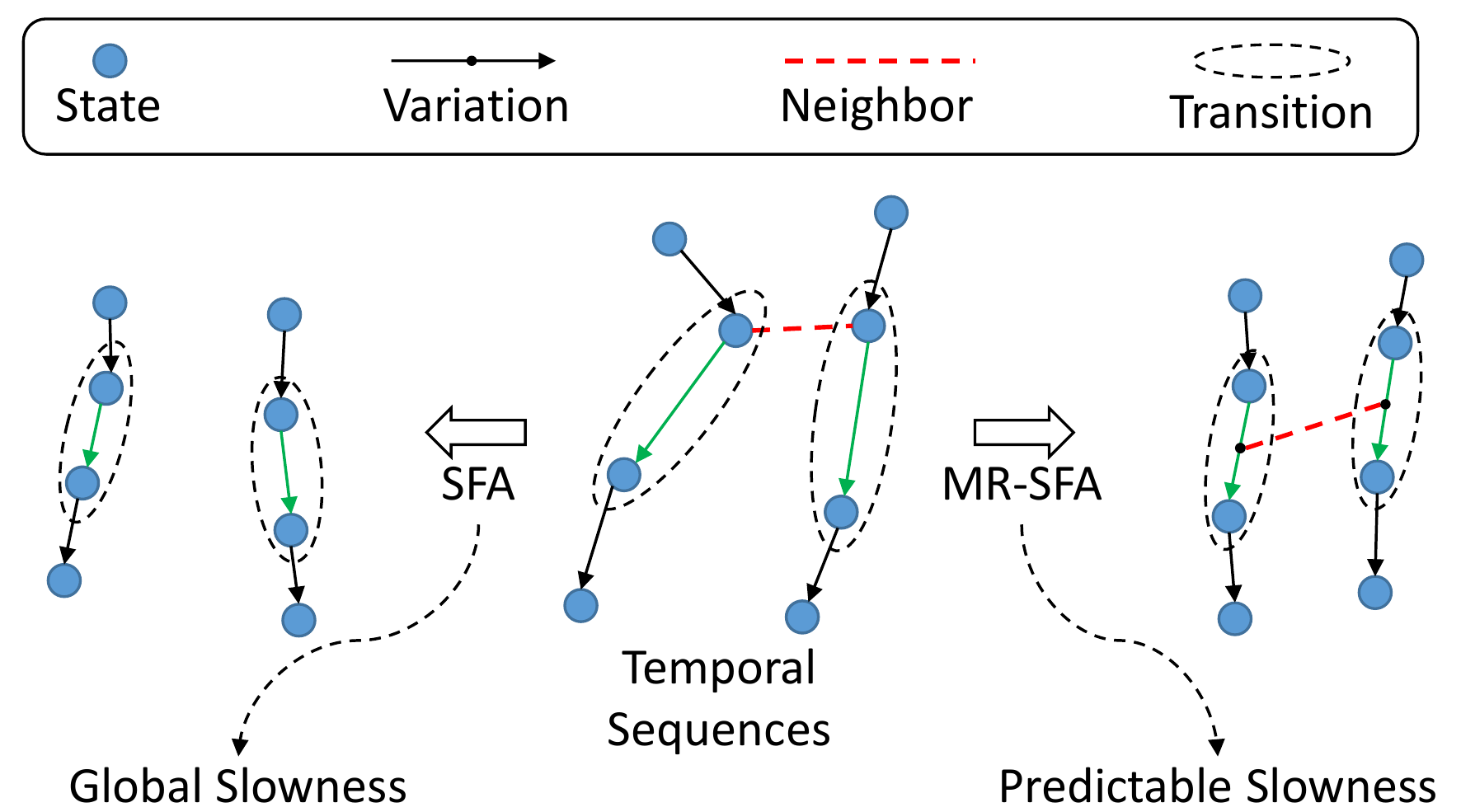}
	\caption{A conceptual illustration of the proposed MR-SFA. SFA is illustrated for a comparison.}
	\label{fig_concept}
\end{figure}

More specifically, each state transition in a temporal sequence consists of three elements, i.e., the initial state, the temporal variation, and the final state. In particular, the final state can be determined by the initial state and the temporal transition. 
It has been proven that a dynamic texture can be regarded as a stationary process, and described by linear dynamical systems (LDS) \cite{LDS}.
Although we cannot model temporal transitions accurately by SFA, we can utilize successive temporal states, which can be reliable and predictable if we learn them properly.
Ideally, in the long term, if two transitions have similar initial states, then they should have similar variations. In other word, each temporal variation should be partly predicted by its initial state.
To accomplish this goal, we construct a neighbor relationship of all temporal transitions by their initial states, and constrain the locality of their variations in the learned feature space. 
A conceptual illustration of the proposed MR-SFA is shown in Fig.~\ref{fig_concept}.

Notably, it is essential to construct the neighbor relationship by states, and constrain the variations. For each temporal transition, similar states always result in similar variations. 
However, similar variations might be caused by totally different transitions. 
Therefore, although the constraint is imposed on the variations, the similarity of the temporal transitions should be determined by the initial states.

Defining each temporal transition as a tuple, MR-SFA can be concluded by two aspects: minimizing intra-variations of each tuple and preserving the locality of similar tuples.
Therefore, the initial objective function of MR-SFA is formulated as
\begin{equation}
\begin{aligned}
	& \arg \min_U \left( \sum_i^{t-1} ||\mathbf{\dot{y}}_i||_2^2 + \lambda \sum_i^{t-1} \sum_j^{t-1} S_{ij} ||\mathbf{\dot{y}}_i - \mathbf{\dot{y}}_j||_2^2 \right) \\
	& \textit{s.t.} \indent U^T U = I,
\end{aligned}
\end{equation}
where $S$ is the similarity matrix, and $\lambda$ is a hyper-parameter to balance the weight between the temporal slowness and the regularization. 
The first part of this objective function is identical to SFA, and the second part is the manifold regularization that retains the locality of the temporal transitions.
The similarity matrix $S$ is determined by the initial states of each temporal transition.
Specifically, if $\mathbf{x}_i$ is among the $k$-nearest neighbors of $\mathbf{x}_j$, or $\mathbf{x}_j$ is among the $k$-nearest neighbors of $\mathbf{x}_i$, we set
\begin{eqnarray}
	S_{ij} = \exp (- \frac {||\mathbf{x}_i - \mathbf{x}_j||^2} {r}),
\end{eqnarray}
and $S_{ij} = 0$ otherwise. Here, $r$ is a hyper-parameter that regulates the weight of the neighboring connections. 
The objective function incurs a heavy penalty if the temporal variation of similar transitions are mapped far apart. 
In this way, the locality of the temporal variations in similar transitions is preserved, and the variations can be partly predicated by their current states.

Following some simple derivations, we then have
\begin{eqnarray}
\begin{aligned}
	& \sum_i^{t-1} ||\mathbf{\dot{y}}_i||_2^2 + \lambda \sum_i^{t-1} \sum_j^{t-1} S_{ij} ||\mathbf{\dot{y}}_i - \mathbf{\dot{y}}_j||_2^2 \\
	& = tr(\dot{Y} \dot{Y}^T) + \lambda tr(\dot{Y} (D-S) \dot{Y}^T) \\
	& = tr(\dot{Y} (I + \lambda (D - S)) \dot{Y}^T) \\
	& = tr(\dot{Y} L \dot{Y}^T), 
\end{aligned}
\end{eqnarray}
where $D$ is a diagonal matrix with entries $D_{ii} = \sum_j S_{ij}$, and 
\begin{equation}
	L = I + \lambda (D - S).
\end{equation}
Thus far, the initial objective function can be reformulated as
\begin{equation}
	\arg \min_U tr(\dot{Y} L \dot{Y}^T) \indent \textit{s.t.} \indent U^T U = I.
\end{equation}

The matrix $D$ provides a measurement of the importance of each tuple. If a tuple has more neighbor tuples, then it might be more predictable.
Therefore, we add an additional constraint $\dot{Y}^T D \dot{Y} = I$ as the weight of each tuple.
The new object function is formulated as 
\begin{equation}
\begin{aligned}
	& \arg \min_U tr(\dot{Y} L \dot{Y}^T) \\
	& \textit{s.t.} \indent \dot{Y}^T D \dot{Y} = I \indent \textit{and} \indent U^T U = I.
\end{aligned}
\end{equation}
Notably, different from the constraint $Y^T Y = I$ used in standard SFA, the new constraint $\dot{Y}^T D \dot{Y} = I$ cannot guarantee that the learned new representation has an identity covariance matrix.

To eliminate the constraint $\dot{Y}^T D \dot{Y} = I$, the objective function can be further reformulated as 
\begin{equation}
	\arg \min_U \frac {tr(\dot{Y} L \dot{Y}^T)} {tr(\dot{Y} D \dot{Y}^T)} \indent \textit{s.t.} \indent U^T U = I.
\end{equation}
Considering that $Y=U^T X$, we have
\begin{equation}
	\arg \min_U \frac {tr(U^T (\dot{X} L \dot{X}^T) U)} {tr(U^T (\dot{X} D \dot{X}^T) U)} \indent \textit{s.t.} \indent U^T U = I.
\end{equation}
Last, the solution $U$ can be obtained by the solution of the generalized eigenvalue problem
\begin{equation}
	(\dot{X} L \dot{X}^T)U = \Lambda (\dot{X} D \dot{X}^T) U, 
\end{equation}
where $\Lambda$ is a diagonal matrix of eigenvalues.
In practice, the first solution or few solutions that correspond to eigenvalues that are close to zero might be caused by noise. 
These noisy solutions should be abandoned, and the remaining solutions can be used for subsequent processing.

Although Sprekeler \cite{SFA_LE} showed that SFA is related to Laplacian eigenmaps \cite{Laplacian} for encoding the locality of the neighboring samples, the proposed MR-SFA focuses on variations in temporal transitions.
The largest advantage of temporal slowness is to utilize the natural temporal relationship, which is stronger than the relationship constructed by k-nearest neighbors in the original space. 
Successive states in a sequence might be very different due to the complex temporal variation. MR-SFA resolves these variations despite the locality in the original space.

Moreover, the aforementioned algorithm uses only one temporal sequence for learning. It can be simply extended to more sequences by evaluating all possible temporal transitions as tuples.
Overall, MR-SFA can be summarized as in Algorithm~\ref{alg_mrsfa}.

\begin{algorithm}
\caption{MR-SFA}
\label{alg_mrsfa}
\begin{algorithmic}[1]
	\Require A temporal sequence $X = [\mathbf{x}_1, \cdots, \mathbf{x}_t] \in \mathbb{R}^{p \times t}$, where $\mathbf{x}_i$ is a column vector that indicates the state of sequence at time $i$. Here $\{\mathbf{x}_i\}$ is assumed to be centered and whitened.
	\Ensure A slowly varying and partly predicable representation $Y=U^T X \in \mathbb{R}^{q \times t}$, and the projection matrix $U \in \mathbb{R}^{p \times q}$.
	\State Construct the similarity matrix $S$ by the k-nearest neighbor of $\{ \mathbf{x}_i \}$.
	\State $D_{ii} \gets \sum_j S_{ij}$, and $L \gets I + \lambda (D - S)$.
	\State $A \gets \dot{X} L \dot{X}^T $
	\State $B \gets \dot{X} D \dot{X}^T $
	\State Solve the generalized eigen-decomposition problem $AU=\Lambda BU$ to obtain the solution $U$ and $Y$.
\end{algorithmic}
\end{algorithm}

\section{MR-SFA for Dynamic Texture Recognition}
\label{sec_mrsfa_features}

\begin{figure*}
	\centering
	\includegraphics[width=7in]{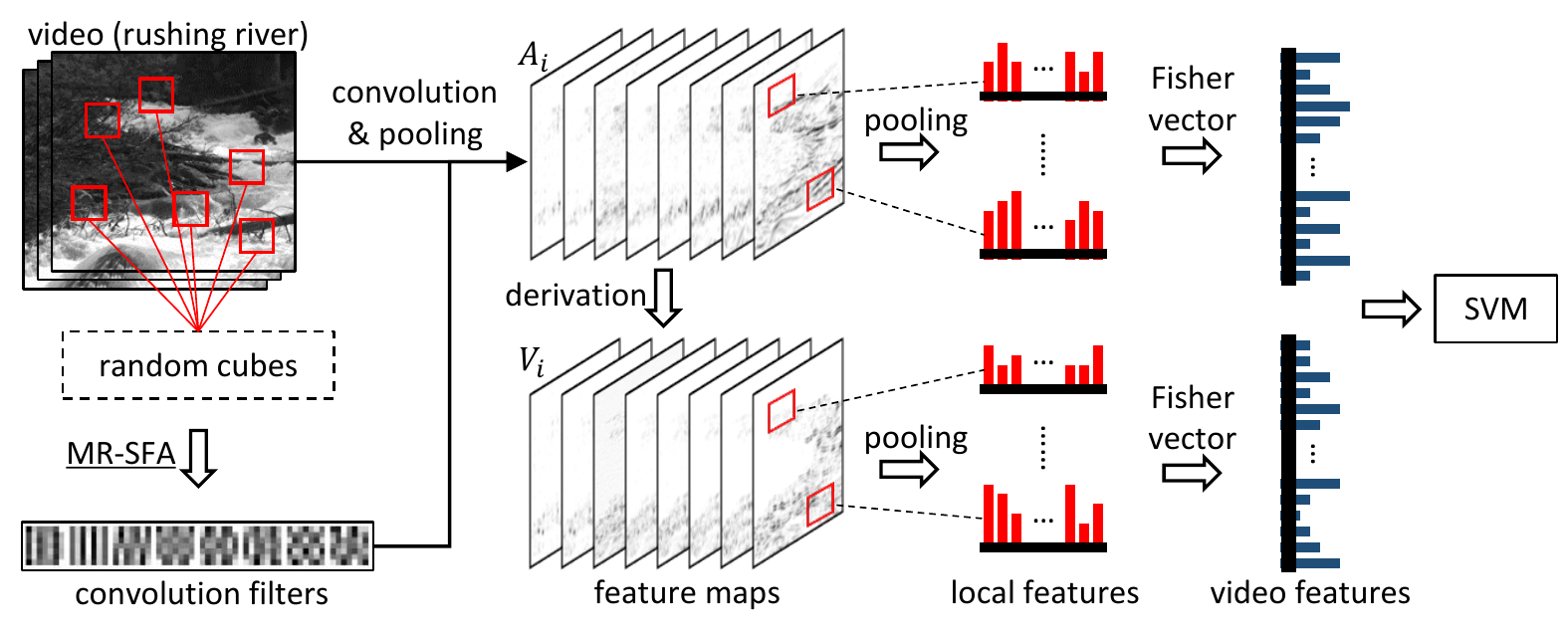}
	\caption{An illustration of the proposed dynamic texture recognition framework.}
	\label{fig_pipeline}
\end{figure*}

This section presents the proposed feature extraction process for dynamic texture recognition.
We learn feature extraction functions from randomly extracted small video cubes, and we use them as convolution filters to generate feature maps. 
Then, spatial and temporal pooling are employed to extract local features from feature maps. 
Last, all of the extracted features are encoded by Fisher vectors to obtain a video-level representation for classification.
An illustration of the proposed framework is shown in Fig.~\ref{fig_pipeline}.

\subsection{Learning Convolution Filters}

Generally speaking, larger receptive fields contain more high-level information. However, it is impractical to obtain a high-level semantic representation simply by a linear projection.
We choose to learn feature extraction functions for small receptively fields (e.g. a local volume of spatial-temporal size $7{\times}7{\times}15$).
Two pre-processing procedures are required for applying MR-SFA. 
In practice, thousands of small video sequences are used for learning.
We denote the size of each sequence as $h_s{\times}w_s{\times}l_s$, where $h_s$, $w_s$, and $l_s$ are the height, width and length respectively.
First, frame-based sequences are reformatted into cube-based sequences to obtain long-term stable temporal transitions. This procedure is similar to the reformatting that is proposed in \cite{SFA_Action}.
Specifically, we reformat each sequence into a new sequence that consists of elemental cubes of size $h_s{\times}w_s{\times}d_s$, where $d_s$ is the length of each elemental cube. 
The number of elemental cubes in each reformatted sequence is $l_n=l_s-d_s+1$.
In addition to achieving long-term temporal slowness, the reformation procedure enables reliable temporal prediction. 
Secondly, principal component analysis (PCA) and whitening are employed to reduce the dimension of all elemental cubes from $h_s{\times}w_s{\times}d_s$ to $m$.
After applying PCA whitening, the size of each sequence is $m{\times}l_n$.
Last, MR-SFA is applied to learn feature extraction functions from these sequences.

\begin{figure}[!t]
	\centering
	\subfloat[Convolution filters learned by MR-SFA.]{\includegraphics[width=2.5in]{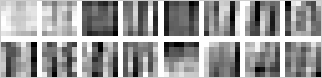}}
	\hfil
	\subfloat[Convolution filters learned by standard SFA]{\includegraphics[width=2.5in]{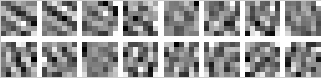}}
	\caption{The visualization of 16 slim convolution filters. Here, 16 filters are shown from top-left to bottom-right.}
	\label{fig_filters}
\end{figure}

Combining PCA and MR-SFA, features can be extracted directly from raw videos. 
Specifically, we denote projection matrices of MR-SFA and PCA whitening as $U$ and $P$, and the mean of all training samples as $\tilde{\mathbf{b}}$. 
The feature extraction function is formulated as
\begin{equation}
	g(x) = U^T P^T (\mathbf{x} - \tilde{\mathbf{b}}) = W^T \mathbf{x} + \mathbf{b},
\end{equation}
where 
\begin{equation}
	W = (U^T P^T)^T = PU
\end{equation}
represents the weights (i.e. convolution filters), and 
\begin{equation}
	\mathbf{b} = - W^T \tilde{\mathbf{b}}
\end{equation}
represents the biases. Therefore, the feature extraction can be performed simply by applying this liner function.
Each column of $W = [{\mathbf{w}}_1, \cdots, {\mathbf{w}}_q]$ is a 3D convolution filter of size $h_s{\times}w_s{\times}d_s$.
All of the slices in a learned filter are similar to one another. Therefore, we use slim filters instead of full-length 3D filters \cite{TVA}.
Specifically, the first frame of each filter is used to replace the original full-length 3D filters.
In this way, the size of each filter is reduced from $h_s{\times}w_s{\times}d_s$ to $h_s{\times}w_s$.
We denote these slim filters as $\hat{W} = [\hat{\mathbf{w}}_1, \cdots, \hat{\mathbf{w}}_q]$.
The convolution can be performed more efficiently with these 2D filters.
A visualization of learned convolution filters $\{\hat{w}_i\}$ is shown in Fig.~\ref{fig_filters}.
As shown in the figure, filters learned by standard SFA are noisy due to the complex temporal variations in dynamic textures. Filters learned by MR-SFA are more reliable compared with filters learned by standard SFA.

\subsection{Feature Maps}

Feature maps are obtained by convolution and pooling. 
We denote each video as $[I_1, \cdots, I_n]$, where $I_i$ is the $i$-th frame, and $n$ is the total number of frames.
The $j$-th convolution output map of frame $I_i$ is obtained by
\begin{equation}
	M_i^{(j)} = \hat{\mathbf{w}}_j \otimes I_i + b_j,
\end{equation}
where the $\otimes$ operator indicates the convolution operation, and $\hat{\mathbf{w}}_j$ and $b_j$ are the $j$-th convolution filter and bias respectively.
We further perform a spatial pooling to reduce the spatial size of $M$. The new output is denoted as 
\begin{equation}
	\hat{M}_i^{(j)} = h(g(M_i^{(j)})),
\end{equation}
where $h(\cdot)$ and $g(\cdot)$ are the spatial pooling and activation function respectively.
By default, we simply use an absolute value function as the activation function, i.e. $g(x) = |x|$.
The choosing of the activation function $g(\cdot)$ will significantly affect the recognition results.
Both max-pooling and average-pooling can be used as the spatial pooling operation $h(\cdot)$. 
In our work, we use a non-overlapped max pooling of size $2{\times}2$ or $4{\times}4$.

\begin{figure}
	\centering
	\includegraphics[width=3.4in]{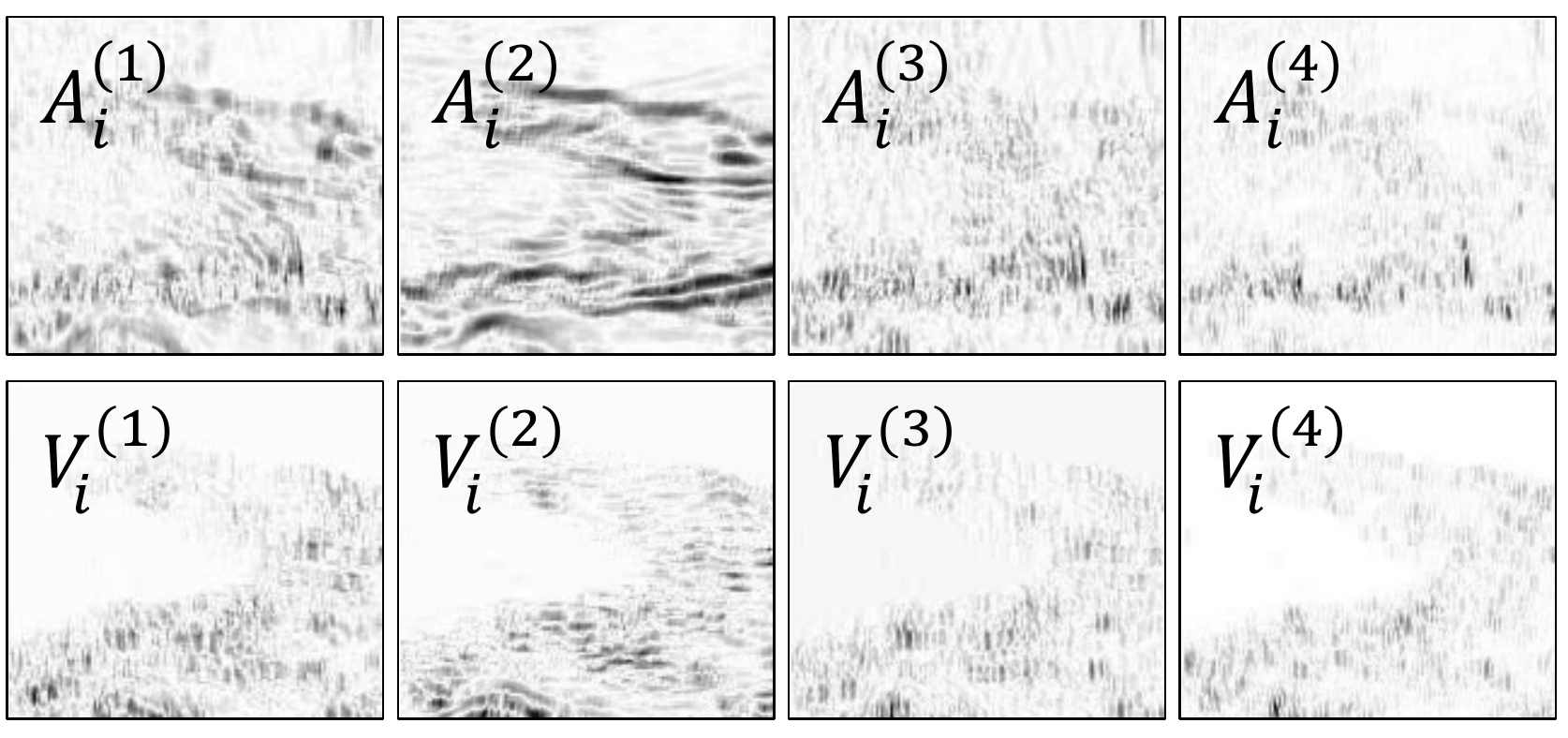}
	\caption{The visualization of feature maps generated by the first four convolution filters. Here, the original frame of these feature maps is the input video frame (rushing river), which is shown in Fig.~\ref{fig_pipeline}.}
	\label{fig_feature_maps}
\end{figure}

Two types of feature maps are obtained from $\hat{M}$ for the subsequent feature extraction. Specifically, the $j$-th appearance feature map of a frame $I_i$ is obtained by
\begin{equation}
	A_i^{(j)} = \left| \hat{M}_i^{(j)} \right|,
\end{equation}
where the $|\cdot|$ operator is a element-wise absolute value function.
Appearance feature maps $\{A_i\}_{1 \leq i \leq n}$ can keep tracks of appearance information, which is important for dynamic texture recognition. 
For standard SFA, features that were extracted from appearance feature maps are used for the final representation.
They represent near-static information, which is appearance information, and they are invariant to spatial-temporal variations.
Besides slowly varying features, we also propose variation feature maps for the variation itself.
The $j$-th variation feature maps of a frame $I_i$ are obtained by
\begin{equation}
	V_i^{(j)} = \left| \hat{M}_i^{(j)} - \hat{M}_{i+1}^{(j)} \right|.
\end{equation}
Variation feature maps $\{V_i\}_{1 \leq i \leq n-1}$ carry well distributed temporal variation information on dynamic textures.
Features extracted from variation feature maps significantly improve the representation of dynamic textures.
An illustration of some of the extracted feature maps is shown in Fig~\ref{fig_feature_maps}.

\subsection{Local Features}

\begin{figure}
	\centering
	\includegraphics[width=3.4in]{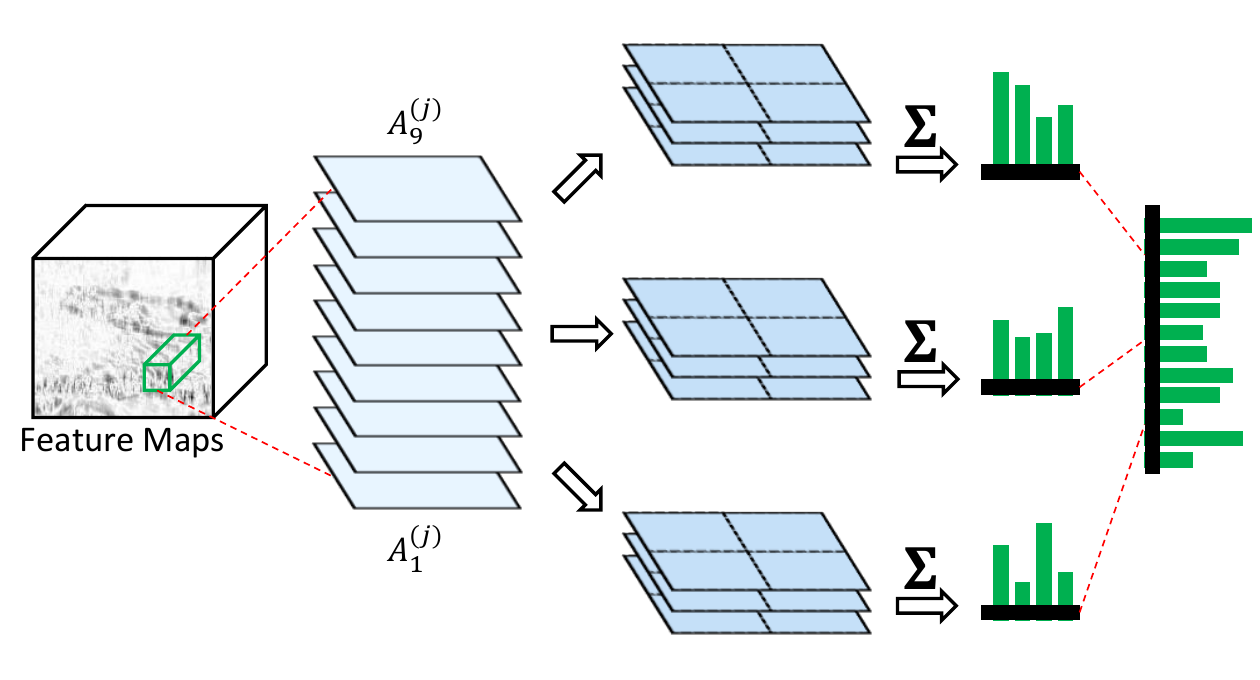}
	\caption{An illustration of spatial-temporal pooling in a sequence of feature maps.}
	\label{fig_pooling}
\end{figure}

After the feature maps are obtained, a spatial-temporal pooling is performed to accomplish local feature extraction.
Appearance and variation features are extracted independently due to the differences between the appearance and variation feature maps. 
Therefore, each set of convolution filters result in two sets of local features in our approach. 
Considering a dynamic texture as a multi-channel video cube (each channel is a feature map), each local feature is extracted by pooling a local volume of size $h_p{\times}w_p{\times}l_p$, where $h_p$, $w_p$, and $l_p$ are the height, width and length of the local volume, respectively.
The spatial and temporal stride of the local feature extraction is denoted as $s_s$ and $s_t$.
An illustration of the pooling procedure in a sequence of feature maps is shown in Fig.~\ref{fig_pooling}.
In our work, average pooling is used, it maintains more valuable information compared with max pooling.

First, spatial pooling is performed.
For each slice of size $h_p{\times}w_p$ in a local volume, pooling is performed in four equally divided sub-regions (i.e., top-left, top-right, bottom-left, bottom-right) of size $\frac{h_p}{2} \times \frac{w_p}{2}$.
Considering that we have eight feature maps that were generated by eight convolution filters, each slice in the local volume can be described by a $4{\times}8$-dimensional feature.
Furthermore, normalization is applied to each slice by dividing their L2-norms, to obtain better generalization.

Second, temporal pooling is performed on an entire local volume.
Slices in a local volume are equally divided into three parts of length $\frac{l_p}{3}$, and then, they are pooled together. These three parts are concatenated as the final representation of the local volume.
Thus far, each local volume is described by a local feature of size $3{\times}4{\times}8$. 

\subsection{Video Representation}

A large number of local features is extracted from each video. Each feature is a descriptor for a local volume.
A bag-of-words model is employed to encode all of the local features into a high-level representation. 
Here we use the Fisher vector for the feature encoding procedure \cite{FV}. The Fisher vector encodes low-level features by their first- and second-order statistics.
Due to the orthogonal natural of the learned filters, we divide learned filters into smaller sets, and each set consists of eight filters.
Following the group of filters, the feature maps are also divided into different groups. The feature extraction is performed independently in each group. 
In this way, the complexity of each feature set is further reduced. The features can be well described by a small Gaussian mixture model (GMM) for Fisher vectors. Using more filters in a set will result in the under-fitting of GMM, while using fewer filters in a set will result in a bad feature description.

At this point, the dimensionality of each set of local features is $3{\times}4{\times}8=96$. A PCA whitening is performed to reduce its dimensionality to 48 for encoding.
Moreover, we also introduce a multi-scale feature extraction. Specifically, local features are extracted from different spatial scales, and then, they are encoded together by Fisher vectors. 
In practice, two or four extra spatial scales are sufficient.
After applying Fisher vectors, a power normalization and an L2-normalization are applied to each set of encoded features.
Then, all of them are concatenated as the final representation of each dynamic texture, and an extra L2-normalization is further applied.
Last, a one-against-all linear support vector machine (SVM) is employed for classification.

\section{Experiments}
\label{sec_experiments}

\begin{figure*}
	\centering
	\includegraphics[width=6in]{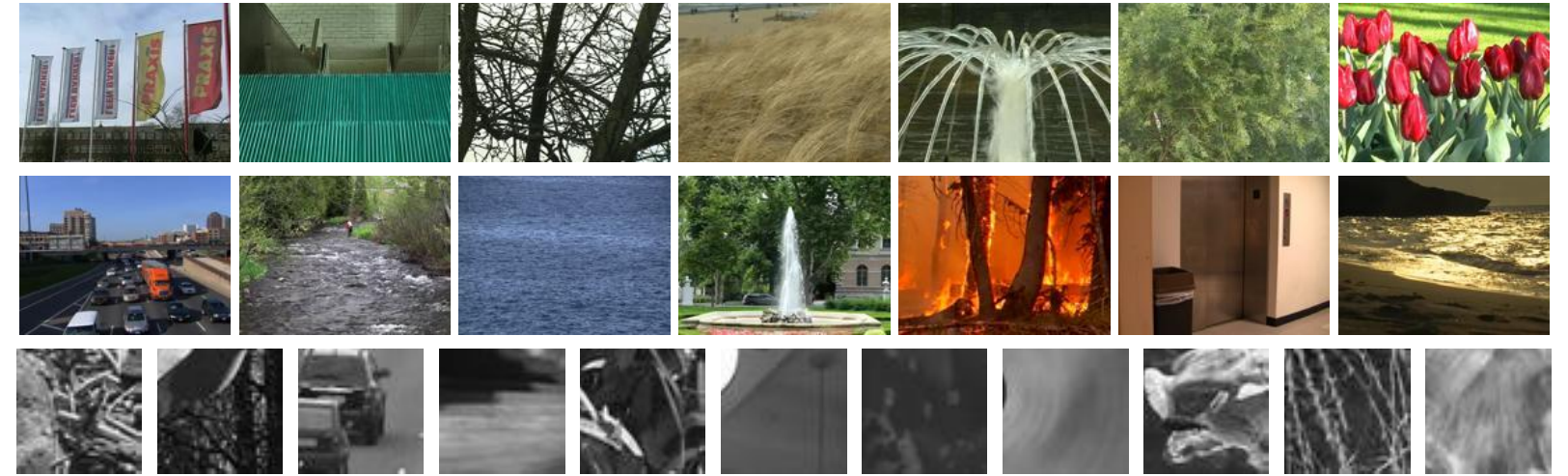}
	\caption{Representative frames of datasets used in our experiments; the rows from top to bottom are Dyntex, YUPENN and Dyntex++ respectively.}
	\label{fig_datasets}
\end{figure*}

The proposed approach was evaluated on three datasets, which range from dynamic texture recognition to dynamic scene recognition. Some frames extracted from the used datasets are shown in Fig.~\ref{fig_datasets}.

The \textbf{DynTex} dataset \cite{Dyntex} is a dynamic texture dataset that consists of more than 650 high-quality videos, e.g., sea grass, trees, smoke, escalator and traffic.
We use the down-sampled version, which all videos are resized to the resolution of $352{\times}288$. The standard classification benchmark divides DynTex into 3 subsets. The \textbf{Alpha} dataset consists of 3 classes with 60 videos, the \textbf{Beta} dataset consists of 10 classes with 162 videos, and the \textbf{Gamma} dataset consists of 10 classes with 275 videos. 
We use videos from the Alpha dataset for the cross-validation, and we report the results on the Beta dataset and Gamma dataset.
In particular, we follow the standard evaluation protocol and report the mean accuracy of the leave-one-video-out (LOO) cross-validation \cite{Dyntex}.
Moreover, we also follow a recently proposed alternative evaluation protocol \cite{TensorDictionary}. 
Specifically, five videos from each category are used for training and the remainder are used for testing. We report the mean accuracy on 20 random splits.

The \textbf{YUPENN} dynamic scene dataset \cite{YUPENN} is introduced to emphasize scene specific temporal information instead of camera-induced aspects. All of the videos in the dataset are captured by a stationary camera. 
The dataset consists of 14 dynamic scene categories, and each category contains 30 color videos.
Videos in the dataset contain significant differences in resolution, frame rate, scale, illumination and viewpoint. 
We follow the leave-one-video-out cross-validation protocol and report the average accuracy as the final result \cite{YUPENN}.

The \textbf{DynTex++} dataset \cite{DynTex++} consists of 36 types of dynamic textures. Each type of dynamic texture contains 100 gray videos of size $50{\times}50{\times}50$. 
Following the standard evaluation protocol, we train on half the samples of each category and test on the remaining samples, and we report the average accuracy of 20 random splits as the final result \cite{DynTex++}. 

\subsection{Implementation Details}

In our experiments, all of the parameters were set according to th following descriptions, unless stated otherwise:

We randomly extracted 100,000 small video cubes from 100 videos to learn convolution filters by MR-SFA. 
The size $h_s{\times}w_s{\times}l_s$ of these cube was $7{\times}7{\times}15$.
These frame-based sequences were further reformatted into cube-based sequences. The length of each elemental cube was $d_s=6$, and the number $l_n$ of elemental cubes in each reformatted sequence was 10.
The dimensionality of each elemental cube was reduced to $m=64$ by PCA whitening, and then, MR-SFA was performed to obtain convolution filters. The number $k$ of nearest states was set to 5, the hyper-parameter $r$ was set to $\frac{m}{2}=32$, and the weight $\lambda$ of manifold regularization was set to $\lambda=0.1$.
Twenty four convolution filters of size $7{\times}7$ were learned, and then, they were equally divided into three groups.
Three sets of variation features and three sets of appearance features were obtained for the final representation.
For each set of features, we trained a GMM with 16 clusters for Fisher vectors from 16,000 randomly sampled local features. 

All of the videos were converted to gray videos, and they were truncated to 256 frames.
The convolution was performed densely by a stride of one.
For the DynTex dataset and YUPENN dataset, a non-overlapped max pooling of size $4{\times}4$ was performed after the convolution.
The volume size $h_p{\times}w_p{\times}l_p$ of each local feature was $8{\times}8{\times}15$. The spatial stride $s_s$ was 1, and the temporal stride $s_t$ was 3.
Five spatial scales were used for feature extraction; they were $[1, 0.5^{\frac{1}{2}}, 0.5, 0.5^{\frac{3}{2}}, 0.25]$.
The size of the videos in the DynTex++ dataset was much smaller (i.e. $50{\times}50{\times}50$), thus we performed a non-overlapped max pooling of size $2{\times}2$ after the convolution.
The volume size $h_p{\times}w_p{\times}l_p$ of each local feature was set to $6{\times}6{\times}9$. The spatial stride $s_s$ and the temporal stride $s_t$ was set to 1 and 3, respectively.
Five spatial scales were used, they were $[2, 0.5^{-\frac{1}{2}}, 1, 0.5^{\frac{1}{2}}, 0.5]$.
All of the experiments were implemented by Matlab 2014a on a Linux system, and they were conducted on a server that had two Intel Xeon E5-2650 V1 CPUs and 128G RAM. 

\subsection{Parameter Evaluation}

To evaluate the parameters used in our experiments efficiently, we constructed a subset based on the DynTex++ dataset. 
We randomly chose ten videos from each category for the subset.
There are 360 videos from 36 categories in total. Three videos in each category were used for training, and the remainder were used for testing. 
The average accuracy on 30 random splits was reported as the final result.
To further speed up the evaluation, the stride of the convolution was increased to two.

\begin{figure}
	\centering
	\includegraphics[width=3.4in]{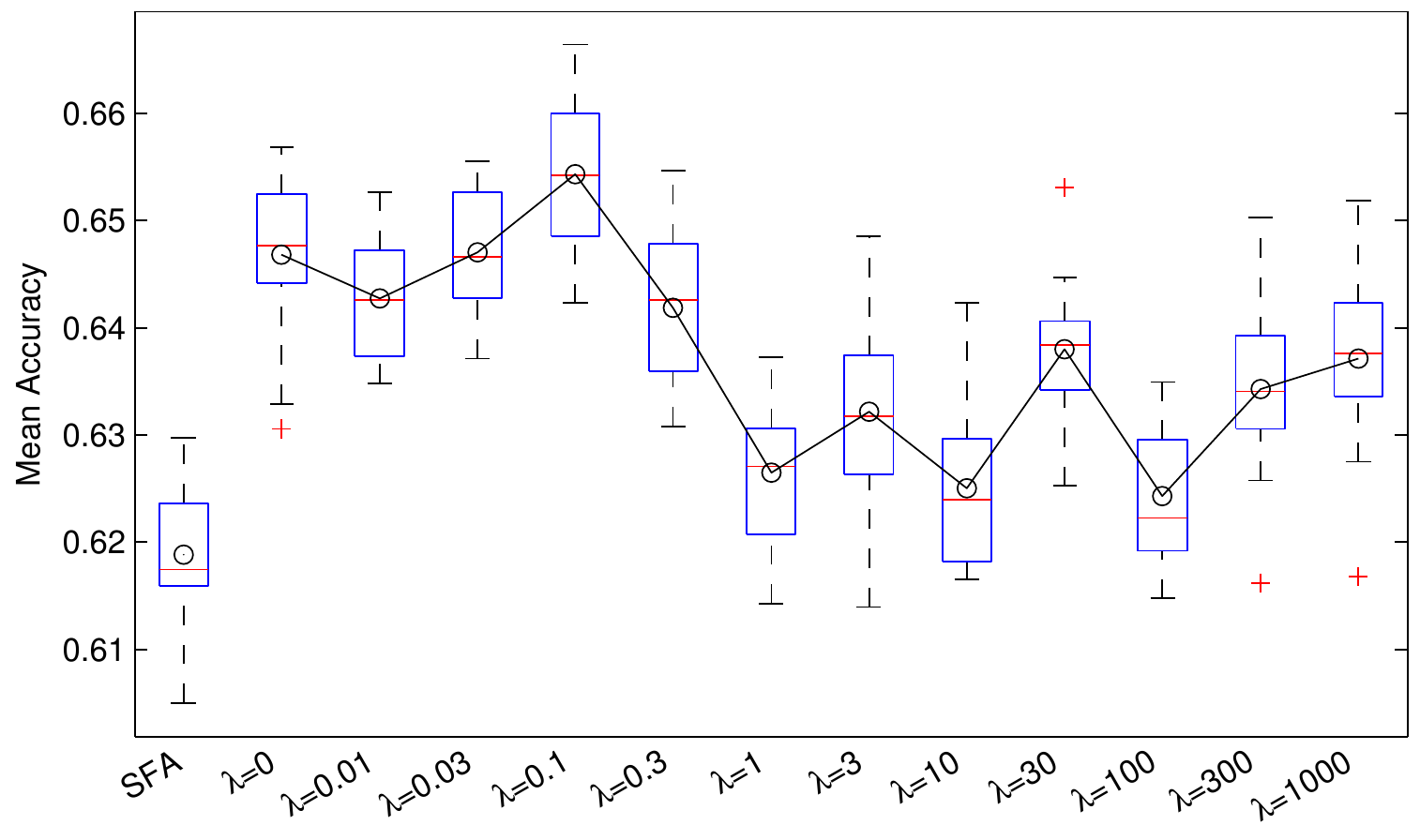}
	\caption{The evaluation of the the parameter $\lambda$. Each value of $\lambda$ was evaluated 20 times. The mean accuracies obtained by different values are marked as circles and connected by a polyline.
		In addition to MR-SFA, the results of standard SFA are also reported as a baseline.}
	\label{fig_lambda}
\end{figure}

First, we analyzed the influence of the parameter $\lambda$, which is the weight of the manifold regularization. 
In addition to $\lambda=0$, values that ranged from 0.01 to 1000 were evaluated. 
The evaluation of each value was repeated 20 times to obtain a credible result.
As shown in Fig.~\ref{fig_lambda}, good results can be achieved by $\lambda \leq 0.3$, and the best result was achieved by $\lambda=0.1$. 
Using a large $\lambda$ will corrupt the temporal slowness of the extracted features, and thus, they performed poorly.
Notably, using $\lambda=0$ also achieved a competitive result due to the additional manifold constraint $\dot{Y}^T D \dot{Y} = I$.
Besides MR-SFA, the results obtained by SFA are also reported as the baseline. In particular, MR-SFA were replaced with SFA, and all of the other parameters were similar.
MR-SFA significantly outperforms standard SFA. 
The improvement comes from two aspects: the regularization for partial prediction, and the weight constraint of the tuples.
Based on this evaluation, we use $\lambda=0.1$ for all subsequent experiments.

\begin{figure}
	\centering
	\includegraphics[width=3.4in]{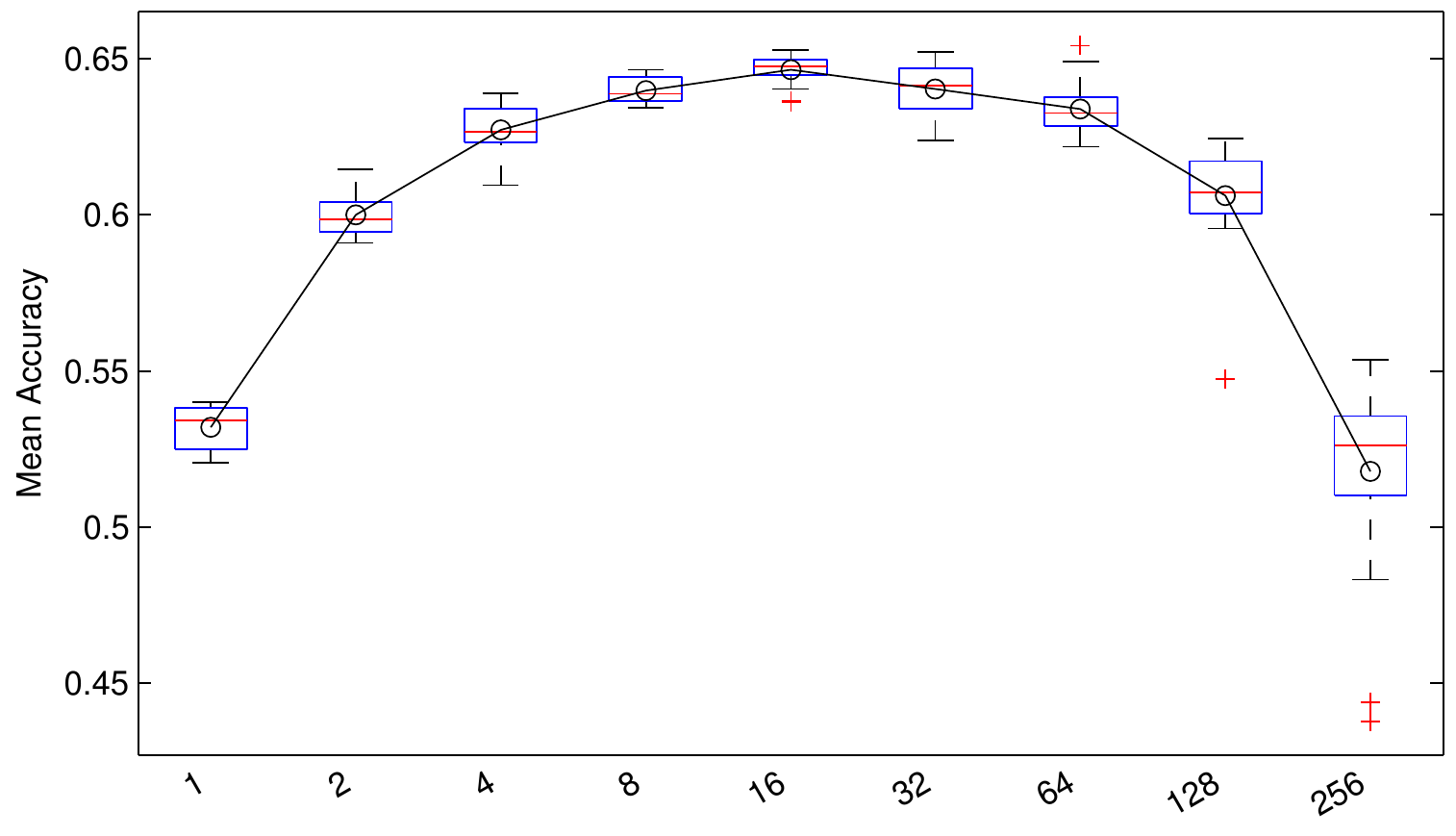}
	\caption{The Evaluation of different GMM clusters for Fisher vectors. Each value was evaluated 20 times. The mean accuracies that were obtained by different values are marked as circles and connected by a polyline.}
	\label{fig_gmm}
\end{figure}

Second, we evaluated the number of GMM clusters that were used for the video representation. 
We tested different GMM cluster numbers that ranged from 1 to 256. Similar to the previous evaluation, the evaluation of each number was repeated 20 times, and the results are shown in Fig.~\ref{fig_gmm}. 
The best result was obtained using 16 GMM clusters. 
The numbers of GMM clusters that are in the range from eight to 64 are competitive compared with the best. 
Notably, results obtained by only one GMM cluster outperform results obtained by 256 GMM clusters.
The proposed dynamic texture recognition relies on a small amount of features. Using a large number of GMM clusters results in overfitting.

\begin{table}[!t]
	\renewcommand{\arraystretch}{1.1}
	\caption{The recognition accuracy (\%) obtained by different activation functions and pooling methods on the DynTex++ subset.}
	\label{tab_pooling}
	\centering
	\begin{tabular}{| p{0.2in}<{\centering} | p{0.35in}<{\centering} | p{0.35in}<{\centering} | p{0.35in}<{\centering} | p{0.35in}<{\centering}|}
		\hline
		& Linear & ReLu & Abs           & Square \\ \hline
		Max & 64.5   & 64.6 & \textbf{64.8} & 62.5   \\ \hline
		Avg & 62.8   & 63.9 & 64.5          & 62.1   \\ \hline
	\end{tabular}
\end{table}

\begin{table*}[!t]
	\renewcommand{\arraystretch}{1.1}
	\caption{The recognition accuracy (\%) obtained by different sets of MR-SFA features.}
	\label{tab_features}
	\centering
	\begin{tabular}{| p{1in}<{\centering} | p{0.6in}<{\centering} | p{0.6in}<{\centering} | p{0.6in}<{\centering} | p{0.6in}<{\centering} | p{0.6in}<{\centering} | p{0.6in}<{\centering} |}
		\hline
		\multirow{2}{*}{} & \multicolumn{2}{c|}{DynTex LOO} & \multicolumn{2}{c|}{DynTex Alternative} & \multirow{2}{*}{YUPENN} & \multirow{2}{*}{DynTex++} \\ \cline{2-5}
		& Beta          & Gamma           & Beta          & Gamma                   &                         &  \\ \hline
		AF1               & 96.7          & 94.6            & 82.5          & 79.4                    & 92.8                    & 95.1                      \\
		AF2               & 95.5          & 94.8            & 86.3          & 78.8                    & 93.9                    & 94.5                      \\
		AF3               & 96.1          & 93.7            & 83.9          & 77.5                    & 95.2                    & 95.3                      \\ \hline
		
		AF1+AF2           & 97.5          & 95.8            & 86.7          & 79.7                    & 96.0                    & 96.0                      \\
		AF1+AF2+AF3 (AF)  & 97.3          & 96.2            & 87.6          & 79.7                    & 96.6                    & 96.7                      \\ \hline
		
		VF1               & 97.7          & 95.8            & 85.4          & 80.1                    & 96.4                    & 93.7                      \\
		VF2               & 98.1          & 94.7            & 85.8          & 77.5                    & 93.9                    & 94.2                      \\
		VF3               & 96.9          & 94.9            & 87.1          & 79.0                    & 94.6                    & 94.3                      \\ \hline
		
		VF1+VF2           & 98.4          & 95.8            & 87.5          & 80.2                    & 96.3                    & 95.2                      \\
		VF1+VF2+VF3 (VF)  & 98.6          & 96.8            & 88.3          & 80.6                    & 96.6                    & 95.9                      \\ \hline
		
		AF1+VF1           & 99.4          & 97.9            & 88.6          & 82.8                    & 97.5                    & 97.0                      \\
		AF2+VF2           & 97.9          & 97.5            & 89.3          & 82.8                    & 96.0                    & 97.1                      \\
		AF3+VF3           & 97.5          & 96.7            & 88.4          & 82.8                    & 97.5                    & 97.2                      \\ \hline
		
		AF1+VF1+AF2+VF2   & \textbf{99.4} & \textbf{98.1}   & 90.2          & \textbf{84.0}           & 97.4                    & 97.4                      \\
		AF+VF             & 99.0          & \textbf{98.1}   & \textbf{90.4} & \textbf{84.0}           & \textbf{97.9}           & \textbf{97.7}             \\ \hline
	\end{tabular}
\end{table*}

We also evaluated different combinations of activation functions $g(\cdot)$ and pooling methods $h(\cdot)$.
Each combination was evaluated 20 times, and the mean results are shown in Table~\ref{tab_pooling}. 
Among all of the combinations, the max pooling and the absolute value function achieved the best performance. The max pooling outperforms the average pooling.
The absolute value function takes advantage of both the positive and negative responses. Thus it performs better than the linear function and the rectified linear unit (ReLu) \cite{RELU} in our experiments. The square function performed poorly compared with the absolute value function. Although it works in a similar way compared with the absolute value function, it corrupts the linearity of the original responses.

\subsection{Feature Evaluation}

We further conducted experiments on different datasets to analyze each set of features. Here, each evaluation is repeated 3 times, and the average result is reported. 
In contrast to the experiments conducted on the subsets of the DynTex++ dataset, the experiments here attempted to achieve the best result. Therefore, the variance of the obtained results is small, and it can be ignored. 
There were 24 convolution filters that were separated into three sets, and six sets of features were generated from them.
Three sets of variation features (VF) obtained from variation feature maps $\{V_i\}$ are denoted as VF1, VF2 and VF3, and three sets of appearance features (AF) obtained from appearance feature maps $\{A_i\}$ are denoted as AF1, AF2 and AF3.

As shown in Table~\ref{tab_features}, the combination of all of the features (AF+VF) performed best, and each single feature performed poorly compared with the best result.
Notably, although the first set of filters is the best solution of MR-SFA compared with the others, sometimes they performed worse compared with the other filters. 
This phenomenon might be caused by the noise that exists in the learned features.
As described in the previous section, the first one or few solutions of MR-SFA should be abandoned due to the noise, and we abandoned only the first solution in all experiments.

Using more sets of filters is helpful.
However, a combination of different types of features is more effective. In our experiments, using more than three sets of filters barely improved the accuracy. 
Notably, the best recognition accuracy can be achieved by only using features that were obtained using 16 filters (AF1+VF1+AF2+VF2) on the DynTex Beta dataset and DynTex Gamma dataset.

Both the appearance and motion information are essential to dynamic texture recognition. It is difficult to tell which contributes more. VF outperforms AF on both the DynTex dataset and YUPENN dataset. All of these datasets have a relatively large resolution and complex background, which makes VF more robust than AF.
However, AF outperforms VF on the DynTex++ dataset. This result might be caused by the simplicity of the videos in this dataset.

\subsection{Comparison with State-of-the-Art Approaches}

In this subsection, we compare the proposed approach with state-of-the-art approaches. We also report results that were obtained by SFA for the comparison.

\begin{table}[!t]
	\renewcommand{\arraystretch}{1.1}
	\caption{The recognition accuracy (\%) obtained on the DynTex dataset compared with state-of-the-art approaches, using the leave-one-video-out protocol.}
	\label{tab_approaches_dyntex_loo}
	\centering
	\begin{tabular}{| p{1.5in}<{\raggedright} | p{0.5in}<{\centering} | p{0.5in}<{\centering} |}
		\hline
		Methods                              & Beta          & Gamma         \\ \hline
		DFS \cite{FractalAnalysis}           & 76.9          & 74.8          \\
		MBSIF-TOP \cite{MultiscaleBinarized} & 90.7          & 91.3          \\
		ELM \cite{ExtremeLearning}           & 93.7          & 88.3          \\
		ST-TCoF \cite{TransferringCNN}       & 98.2          & 98.1          \\ \hline
		SFA                                  & 95.8          & 96.8          \\ \hline
		\textbf{MR-SFA}                      & \textbf{99.0} & \textbf{98.1} \\ \hline
	\end{tabular}
\end{table}

\begin{table}[!t]
	\renewcommand{\arraystretch}{1.1}
	\caption{The recognition accuracy (\%) obtained on the DynTex dataset compared with state-of-the-art approaches, using five videos in each category for training.}
	\label{tab_approaches_dyntex_alt}
	\centering
	\begin{tabular}{| p{1.5in}<{\raggedright} | p{0.5in}<{\centering} | p{0.5in}<{\centering} |}
		\hline
		Methods                     & Beta          & Gamma         \\ \hline
		DFS \cite{FractalAnalysis}  & 76.5          & 74.5          \\
		OTF \cite{OTF}              & 75.4          & 73.5          \\
		LBP-TOP \cite{Wavelet}      & 73.4          & 72.0          \\
		OTD \cite{TensorDictionary} & 76.7          & 74.8          \\ \hline
		SFA                         & 86.2          & 79.2          \\ \hline
		\textbf{MR-SFA}             & \textbf{90.4} & \textbf{84.0} \\ \hline
	\end{tabular}
\end{table}

The comparison on the DynTex dataset is shown in Table~\ref{tab_approaches_dyntex_loo} and \ref{tab_approaches_dyntex_alt}. 
MR-SFA outperforms all of the existing approaches on the DynTex dataset.
MBSIF-TOP can be regarded as an improvement over LBP-TOP; it performs well on the DynTex dataset.
A significant improvement can be achieved based on LDS features using ELM.
The spatial-temporal transferred convolutional neural network feature (ST-TCoF) was proposed using a pre-trained convolutional neural network \cite{TransferringCNN}. With the prior knowledge of more than a million images, ST-TCoF outperforms most of the existing features. The results obtained by MR-SFA are slightly better than the results obtained by ST-TCoF.
The oriented template features (OTF) employ SIFT-like feature descriptors and a powerful global statistical descriptor for texture description \cite{OTF}. 
The orthogonal tensor dictionary (OTD) employs tensor-based sparse coding as a dynamic texture descriptor \cite{TensorDictionary}.
MR-SFA outperforms all of these approaches on both evaluation protocols.

\begin{table}[!t]
	\renewcommand{\arraystretch}{1.1}
	\caption{The recognition accuracy (\%) obtained on the YUPENN dataset compared with state-of-the-art approaches.}
	\label{tab_approaches_yupenn}
	\centering
	\begin{tabular}{| p{1.5in}<{\raggedright} | p{0.5in}<{\centering} |}
		\hline
		Methods                        & Accuracy      \\ \hline
		Gabor+SFA \cite{SFA_Scene}     & 85.0          \\
		BoSE \cite{BagsEnergies}       & 96.2          \\
		AlexNet \cite{C3D,AlexNet}     & 96.7          \\
		C3D \cite{C3D}                 & 98.1          \\
		SA-CNN \cite{SA_CNN}           & 98.3          \\
		ST-TCoF \cite{TransferringCNN} & 99.1          \\ \hline
		SFA                            & 97.4          \\ \hline
		\textbf{MR-SFA}                & \textbf{97.9} \\ \hline
	\end{tabular}
\end{table}

The comparison on the YUPENN dataset is shown in Table~\ref{tab_approaches_yupenn}. 
MR-SFA outperforms most of the existing approaches on the YUPENN dataset.
An approach called bags of spacetime energies (BoSE) was proposed for dynamic scene recognition \cite{BagsEnergies}. This approach uses oriented 3D Gaussian third-derivative filters for feature extraction. 
The result obtained by the AlexNet is also reported as a baseline for all of the CNN-based approaches \cite{C3D,AlexNet}. 
The convolution 3D (C3D) \cite{C3D}, the statistical aggregation convolutional neural network (SA-CNN) \cite{SA_CNN}, and the ST-TCoF are CNN-based approaches that involve pre-training on enormous amounts of data.
Compared with these CNN-based approaches, the results obtained by MR-SFA are still competitive.

\begin{table}[!t]
	\renewcommand{\arraystretch}{1.1}
	\caption{The recognition accuracy (\%) on the DynTex++ dataset compared with state-of-the-art approaches.}
	\label{tab_approaches_dyntex_pp}
	\centering
	\begin{tabular}{| p{1.5in}<{\raggedright} | p{0.5in}<{\centering} |}
		\hline
		Methods                               & Accuracy      \\ \hline
		LBP-TOP \cite{MultiscaleBinarized}    & 89.5          \\
		DFS \cite{FractalAnalysis}            & 91.7          \\
		DNG \cite{DirectionalNumber}          & 93.8          \\
		OTD \cite{TensorDictionary}           & 94.7          \\
		Chi-Square LBP-TOP \cite{Chi-Squared} & 97.0          \\
		MBSIF-TOP \cite{MultiscaleBinarized}  & 97.2          \\ \hline
		SFA                                   & 97.0          \\ \hline
		\textbf{MR-SFA}                       & \textbf{97.7} \\ \hline
	\end{tabular}
\end{table}

The comparison on the DynTex++ dataset is shown in Table~\ref{tab_approaches_dyntex_pp}. 
Videos of the DynTex++ dataset have less backgrounds compared with the other datasets. 
Therefore, LBP-TOP and its improvements show significant advantages on the DynTex++ dataset.
Similar to LBP-TOP, DNG extracts features from nine different planes in the video cube.
The chi-squared LBP-TOP was proposed using a chi-squared transformation to better fit the Gaussian distribution \cite{Chi-Squared}.
MR-SFA outperforms all of the state-of-the-art approaches on the DynTex++ dataset.

Overall, both SFA and MR-SFA can achieve competitive results. More specifically, state-of-the-art results on the DynTex dataset and the DynTex++ dataset can be achieved by MR-SFA. 
MR-SFA can obtain significant improvements on both DynTex dataset and DynTex++ dataset compared with standard SFA. The improvements arise from the proposed manifold regularization and the variation features.
Because the YUPENN dataset contains fewer complex temporal transitions, improvements on the YUPENN dataset are relatively small.
Compared with LDS features, the features that were extracted by MR-SFA are well distributed. They can be easily modeled by a small number of GMM clusters for the video representation. In contrast, the parameters of LDS are highly nonlinear. They cannot be compared directly with respect to classification, nor are they well modeled by the conventional bag-of-words models to obtain better representations.
MBSIF-TOP performs best among all of the approaches that extract features from orthogonal planes. 
MR-SFA outperforms MBSIF-TOP due to learned slowly varying features and bag-of-words models. In particular, the temporal complexity is well resolved by learned slowly varying features, and the proposed manifold regularization further improves the robustness of the learned features.
CNN-based approaches (i.e., C3D, SA-CNN and ST-TCoF) perform well among all of the dynamic texture approaches. Especially, pre-trained CNN features contain large amounts of high-level semantic information, and thus, they perform best on the YUPENN dataset. 
Compare with CNN-based approaches, MR-SFA uses only a single convolutional layer, and fewer convolution filters.
MR-SFA outperforms CNN-based approaches on the DynTex dataset. Moreover, MR-SFA can be applied to the DynTex++ dataset, which consists of gray videos that have a small resolution and fewer semantic objects. In this situation, CNN-based approaches cannot be applied directly, but MR-SFA is still efficient and effective.

\subsection{Computational Efficiency}

\begin{table}[!t]
	\renewcommand{\arraystretch}{1.1}
	\caption{The feature extraction speed (frame per second) evaluated on a single CPU core.}
	\label{tab_efficiency}
	\centering
	\begin{tabular}{| p{0.6in}<{\centering} | p{0.5in}<{\centering} | p{0.5in}<{\centering} | p{0.5in}<{\centering} |}
		\hline
		& DynTex++ & DynTex & YUEPNN \\ \hline
		8 filters  & 41.6     & 5.1    & 5.2    \\
		16 filters & 26.3     & 4.2    & 3.9    \\
		24 filters & 20.0     & 3.2    & 3.1    \\ \hline
	\end{tabular}
\end{table}

In this subsection, we analyze the efficiency of the proposed approach.
In our implementation, the convolution was implemented by matrix multiplications, and the pooling was implemented by integral images. 
Therefore, the proposed dense feature extraction can be performed efficiently.
We report the average feature extraction speed on each dataset in Table~\ref{tab_efficiency}.
The evaluation was conducted on a single CPU core running at 2.4GHz. As shown in the table, using more convolution filters linearly increases the computational complexity. 
Due to the low resolution of the videos, the feature extraction on the DynTex++ dataset is efficient compared with others.
Most of the computational time of the feature extraction is spent on convolution and pooling.
In practice, the speed can be simply improved by using more CPU cores, or using GPUs for acceleration. 
In our implementation, we simply employ data parallelism to speed up the feature extraction process.

\section{Conclusion}
\label{sec_conclusion}

We have proposed a novel approach for dynamic texture recognition.
Specifically, we learn feature extraction functions by MR-SFA, and employ convolution and pooling for local feature extraction.
Then dynamic textures are represented using bag-of-words models.
To the best of our knowledge, this study is the first research that introduces SFA to dynamic texture recognition.
The proposed MR-SFA further improves standard SFA by exploring the manifold regularization.
In particular, we construct the neighbor relationship of the initial states of each temporal transition, and retain the locality of their variations in the temporal transition.
In this way, the variation in each temporal transition can be partly predicted by its initial state. This approach ensures that learned features can be robust to complex and noisy temporal transitions. 
Overall, the proposed MR-SFA benefits from following three aspects. 
First, learned local features are not only slowly varying but also partly predictable, and thus, the temporal complexity of the dynamic textures can be better resolved. 
Second, local features are densely extracted by convolution and pooling, which further improves the robustness of extracted local features. 
Last, the bag-of-words model approach ensures that the final representation can be invariant to various spatial-temporal translations, viewpoints, scales, and other aspects.
Experimental results show that competitive results can be achieved by the proposed approach. State-of-the-art results can be achieved on the DynTex and DynTex++ dataset.

\ifCLASSOPTIONcaptionsoff
  \newpage
\fi



\bibliographystyle{IEEEtran}
\bibliography{IEEEabrv,ref}

%

%
%
%




\end{document}